\documentclass{article} 
\usepackage{iclr2027_conference}
\usepackage{times}

\usepackage{amsmath,amsfonts,bm}

\def\eqref#1{equation~\ref{#1}}

\def\1{\bm{1}}

\DeclareMathAlphabet{\mathsfit}{\encodingdefault}{\sfdefault}{m}{sl}
\SetMathAlphabet{\mathsfit}{bold}{\encodingdefault}{\sfdefault}{bx}{n}

\usepackage{hyperref}
\usepackage{url}
\usepackage{graphicx}
\usepackage{booktabs}
\usepackage{array}
\usepackage{placeins}
\usepackage{wrapfig}
\usepackage{needspace}
\usepackage{xcolor}
\usepackage{tikz}
\usepackage{amsmath,amssymb}
\usetikzlibrary{arrows.meta,positioning,calc,fit,backgrounds}%
\title{Building World Models from \\Intrinsic Motivation}

\author{Reece Keller$^{1, 2}$, Declan Kutscher$^{4, 3}$, Xaq Pitkow$^{1, 3}$, Aran Nayebi$^{3, 1, 2}$ \\
$^1$Neuroscience Institute, Carnegie Mellon University\\
$^{2}$Robotics Institute, Carnegie Mellon University\\
$^{3}$Machine Learning Department, Carnegie Mellon University\\
$^{4}$Department of Computer Science, University of Maryland, College Park\\
\texttt{\{rdkeller, xpitkow, anayebi\}@andrew.cmu.edu}\\
\texttt{dkutsche@umd.edu}
}

\iclrfinalcopy 
\begin{document}

\maketitle
\begingroup
\renewcommand{\thefootnote}{}
\footnotetext[0]{Project website: \url{https://neuroagents-lab.github.io/ssrl-agent}}
\endgroup
\begin{abstract}
In the absence of extrinsic rewards or task supervision, human behavior is guided by intrinsic motivation—neural signals that drive curiosity, creativity, play, and other essential facets of intelligence. Intrinsic motivation plays an important developmental role in building robust world models that enable generalization and adaptation. Yet existing studies leave open how a broad range of intrinsic motivations interacts with world-model representation learning to support downstream control. This limits our understanding of which intrinsically collected experiences build reusable models. To address these gaps, we integrate 24 intrinsic objectives, including established methods and explicit variants spanning curiosity, entropy, and competency, into a common model-based reinforcement learning framework. We evaluate the world models they build through zero-shot, few-shot, and multi-task control in locomotion, manipulation, and open-ended gaming environments, using three representation-learning strategies: observation reconstruction, latent prediction, and redundancy reduction. Curiosity-driven exploration with observation reconstruction yields the strongest aggregate family profile, although the best individual pairing depends on the experiment. 
In compositional vision tasks, intrinsic motivation derived from both episodic and lifelong novelty---providing a complementary memory to the world model's generic recurrent state---achieves the best zero-shot transfer and few-shot adaptation.  
These findings motivate designing intrinsic motivation and representation learning jointly. Our results demonstrate world models as a powerful tool for measuring the quality of intrinsically motivated experience, and positions intrinsic motivation as a key ingredient for building world models autonomously.

\end{abstract}

\section{Introduction}
Humans and animals demonstrate a remarkable capacity to create and pursue intrinsic goals, adapt to an ever-changing environment, and learn new skills with a little practice. How do we achieve this feat of autonomy? One promising substrate is an internal model of the world, allowing us to reason and plan in imagination using relevant abstractions.
This hypothesis perhaps originated with Kenneth Craik in \textit{The Nature of Explanation} (1943):
\begin{quote}
\textit{``If the organism carries a ‘small-scale model’ of external reality and of its own possible actions within its head, it is able to try out various alternatives, conclude which is the best of them, react to future situations before they arise, utilize the knowledge of past events in dealing with the present and future, and in every way to react in a much fuller, safer, and more competent manner to the emergencies which face it."}
\end{quote}
The kind of model Craik describes is a generative one, allowing the agent to simulate the causal effects of actions. 
\begin{figure*}[t!]
  \centering
  \resizebox{\textwidth}{!}{\input{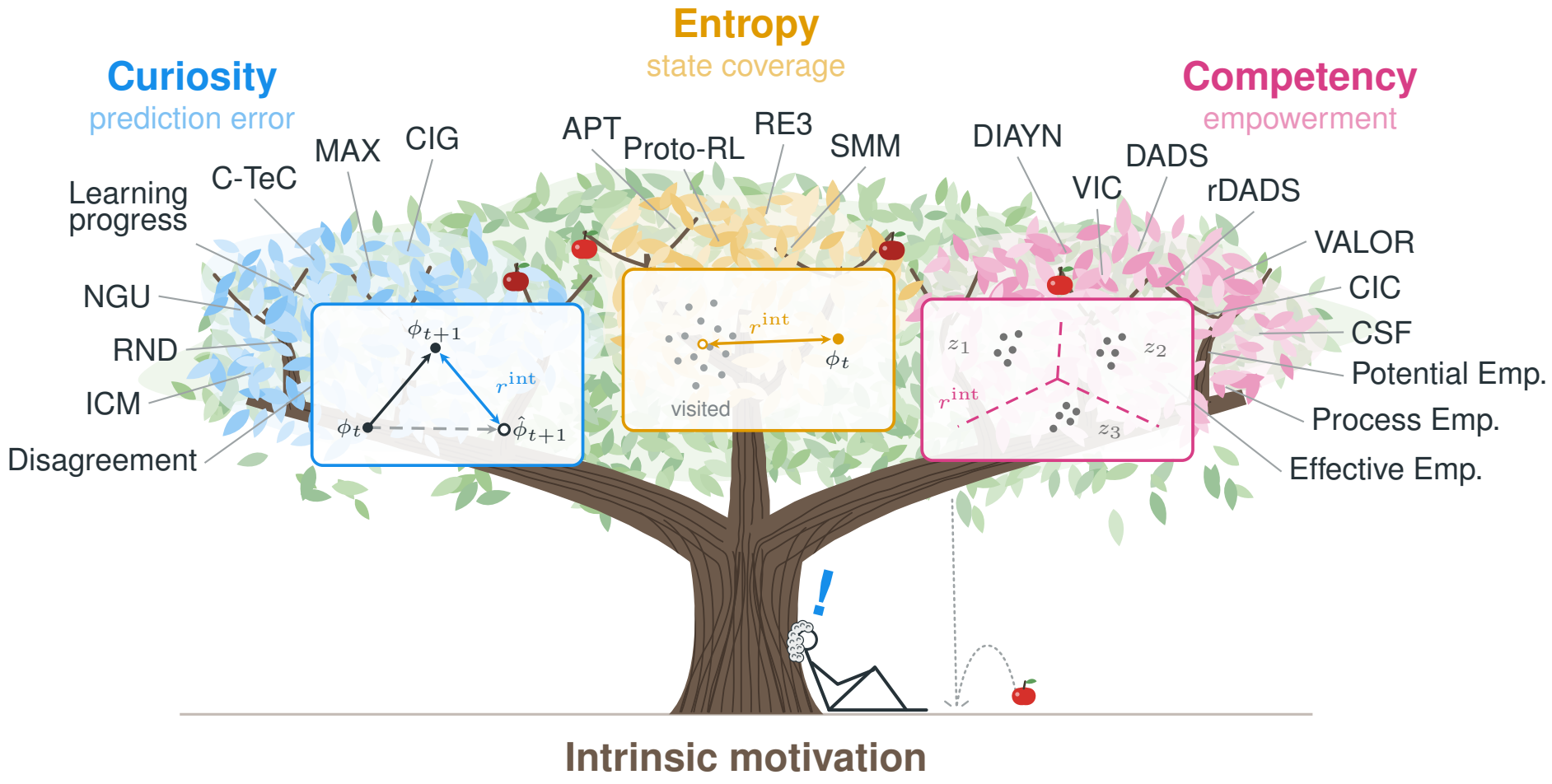}}
  \caption{\textbf{Three families of intrinsic motivation.} Our comparison spans curiosity (blue), entropy (orange), and competency (pink). Leaves illustrate representative objectives and estimator families; insets illustrate prediction error, state novelty, and skill discriminability. Here, $\phi_t$ denotes a state representation, $r^{\mathrm{int}}$ intrinsic reward, and $z_i$ a skill or action.}
  \label{fig:exploration_taxonomy}
\end{figure*}
Such models are a longstanding topic of research in reinforcement learning (RL), culminating in the development of sophisticated model-based algorithms like DreamerV3 and TD-MPC2 that achieve complex behaviors in domains ranging from locomotion to Minecraft \citep{hafner2025dreamerv3,hansen2024tdmpc2}. The RL community typically approaches model-free and model-based algorithms as a methodological choice, and model-free approaches are far more popular by virtue of their simplicity and scalability~\citep{schulman2017proximal,espeholt2018impala}. However, decades of research in neuroscience evidence that human and animal brains build internal predictive models of the world, and that these models are used to plan and guide behavior \citep{tolman1948cognitive,pezzulo2014internally,jensen2026mechanistic,keller2026intrinsic}.
Moreover, recent theoretical work connects sufficiently general goal-directed behavior (including intrinsic goals!) with implicitly learning a world model, under formal assumptions about the goals and environments an agent can solve \citep{richens2025general,nayebi2026capable}. World models are thus a cornerstone of intelligence. Yet still we do not understand how they are refined---beyond their genetic initial conditions---by interaction with the environment over the course of physiological development.
From our perspective, understanding how world models are built up from active experience is a fundamental goal in the study of intelligence.

But in the reinforcement learning framework, the reward function is the only mechanism that shapes behavior. It provides an extrinsic signal for what constitutes success in a given task, allowing agents to learn optimal policies through trial and error. But in real life, and for some of the most interesting and complex behaviors, animals have no such thing. Exploration, play, and creativity---arguably the lifeblood of intelligence---are intrinsically motivated! Naturally, this view raises the question of exactly what it means for an agent to behave intelligently in the absence of an extrinsic task. How do we measure the value of experience? What is intrinsic motivation for?

We propose that intrinsic motivation serves a primary function in developing world models that enable adaptation to new tasks from little to no task-directed experience. Therefore, a window into intelligent, intrinsically motivated behavior is the quality of the world model it builds for downstream control. Previous works on intrinsic motivation in the model-free setting focus on improving state coverage or policy adaptation. These are useful proxy measurements, but neither alone specifies what an agent should learn about its world. The answer we consider here is that experience is valuable insofar as it supports an explicit model that can predict the outcomes of actions and guide future behavior. Studying the interplay between intrinsic motivation and world-model learning therefore offers a way to compare hypotheses about the purpose of exploration, as well as a practical route to better pretraining objectives.

\paragraph{Contributions}
While previous studies investigate intrinsic motivation for learning world models or model-free adaptation, they do not systematically compare how intrinsic objectives and world-model representations jointly shape downstream generalization, only implement a handful of objectives, and use inconsistent RL algorithms \citep{kim2020progress,sekar2020plan2explore,laskin2021urlb,mendonca2021lexa,rajeswar2023mastering,yuan2023euclid}. These issues preclude a clear understanding of where state-of-the-art intrinsic motivation algorithms stand and how they can be improved. 
Our study integrates 24 objectives spanning curiosity, entropy, and competency into a common DreamerV3 framework and compares three representation-learning strategies: observation reconstruction, next-embedding prediction, and current-embedding redundancy reduction. This joint comparison lets us examine how the value of intrinsic experience depends on the representation used for imagination. We evaluate zero-shot control, multi-task transfer, and few-shot adaptation across locomotion, manipulation, and open-ended gaming. For adaptation, we retain the pretrained world model and replay but reinitialize the task heads, distinguishing their transfer benefits from those of retaining a pretrained policy. Appendix~\ref{sec:related} provides a fuller discussion of related work. 

We find that curiosity with observation reconstruction yields the strongest aggregate family profile, although the best individual pairing depends on the environment, observation modality, and evaluation setting. Reconstruction is particularly effective for visual zero-shot control, while reconstruction-free models support strong proprioceptive transfer and competitive few-shot adaptation. Simple curiosity objectives often match or exceed more elaborate variants, suggesting that a strong world-model representation can supply much of the structure needed for intrinsic exploration. We also find that more policy training in imagination can improve initial control yet impair subsequent adaptation in Craftax, exposing a mismatch between extracting behavior from a fixed model and adapting through new experience. This exposes a weakness in Dreamer's ability to maintain a strong persistent memory in abstract, compositional tasks. However, an intrinsic motivation that implements memory at different timescales than the world-model, such as \textit{Never Give Up}, achieves the best results across compositional vision-based experiments. These results motivate designing intrinsic objectives and world-model representations jointly in order to fully benefit from active pretraining.

To summarize, our contributions are:
\begin{itemize}
  \item We integrate 24 intrinsic objectives spanning curiosity, entropy, and competency into a common Dreamer framework with three world-model representation strategies (Figure~\ref{fig:exploration_taxonomy}).
  \item We evaluate zero-shot control, multi-task transfer, and few-shot adaptation across locomotion, manipulation, and open-world gaming environments. 
  \item Our experiments establish \textit{curiosity-driven intrinsic motivation} with observation reconstruction as the strongest pretraining recipe overall. In particular, \textit{Never Give Up}---a multiplicative objective between episodic memory and lifelong novelty---achieves the best zero-shot transfer and adaptation performance in compositional vision-based tasks. 
  \item We identify a scaling mismatch between offline imagination and online adaptation, where more policy training in a pretrained world model impairs subsequent adaptation from task-directed experience.
\end{itemize}


\begin{figure*}[t!]
  \centering
  \includegraphics[width=\textwidth]{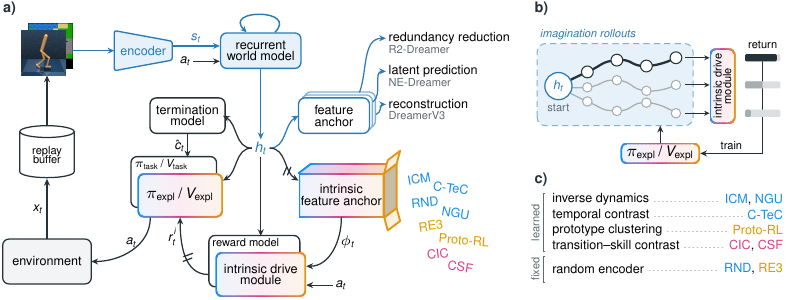}
  \caption{\textbf{Intrinsically motivated Dreamer agents.} (a) Exploration experience trains a shared recurrent world model using reconstruction (DreamerV3), next-embedding prediction (NE-Dreamer), or current-embedding redundancy reduction (R2-Dreamer). The intrinsic drive module (IDM) scores latent states or method-specific intrinsic features. Double slashes denote stopped gradients. (b) Intrinsic rewards on imagined trajectories train the exploration actor--critic. Separate task heads use predicted task rewards for evaluation. (c) Feature spaces used by representative IDMs.}
  \label{fig:overview}
\end{figure*}

\section{Methods}
\label{sec:methods}
Most studies of intrinsic motivation learn representations through model-free policy optimization (e.g., SAC or PPO), making it difficult to separate the effects of the representation and intrinsic objective or assess intrinsic pretraining for zero-shot control and transfer. Thus, we adopt a model-based approach using the basic recipe of Plan2Explore  \citep{sekar2020plan2explore} with DreamerV3's base world model as the common backbone. DreamerV3 has demonstrated remarkable generality across a wide range of tasks~\citep{hafner2025dreamerv3} using a single hyperparameter configuration, making it particularly attractive as a common substrate to unify different intrinsic motivations. The flexibility of Dreamer allows us to test two alternative feature anchors that implement distinct representation learning objectives (Figure~\ref{fig:overview}). We provide a brief background on these algorithms as well as some design choices integrating them with intrinsic motivation. Appendix~\ref{sec:architecture} gives architecture and integration details.

\subsection{Dreamer Backbone and Representation Learning}
Dreamer learns an RSSM from replay and trains behavior through imagined trajectories. Given an  observation $x_t$, the encoder produces $e_t$, and the RSSM combines deterministic memory $h_t$ with a stochastic state $z_t$ to form a world-model state $s_t=(h_t,z_t)$. The posterior conditions on the current observation during training while the prior predicts latent states during imagination. All three backbones retain the same recurrent latent dynamics, continuation prediction, and the Dreamer actor--critic architecture, but differ in how observations anchor the latent state.

DreamerV3 \textit{reconstructs the current observation} with a decoder, balancing the preservation of sensory information against KL regularization that encourages predictable latent representations \citep{hafner2025dreamerv3}. NE-Dreamer predicts the \emph{next encoder embedding} from latent-action history using a causal transformer \citep{bredis2026nedreamer}. R2-Dreamer aligns a linear projection of the \emph{current latent state} with the current encoder embedding \citep{morihira2026r2dreamer}. Both reconstruction-free variants use redundancy reduction to align paired features while preventing collapse \citep{zbontar2021barlow}. Thus, the comparison places sensory reconstruction alongside two forms of latent-space learning, including temporal prediction in the spirit of joint-embedding predictive architectures (JEPA; \citealp{assran2023ijepa}). The specific auxiliary predictors and loss scales also differ between backends, so we emphasize the comparison as one between representation strategies rather than an isolated loss ablation.

We use separate exploration and task actor--critics driven by intrinsic and task reward, respectively. Task rewards can train a detached reward predictor and task actor--critic for evaluation, but their gradients do not update the world model or exploration policy. We report \emph{zero-shot} control before task-directed training interactions. In \emph{few-shot} adaptation, task-directed data and reward update the shared world model, as they do in supervised training.

\subsection{Intrinsic Drive Module (IDM)}
We adapt each intrinsic motivation through an IDM that learns from replay and scores imagined transitions. All IDMs use the deterministic RSSM state component $h_t$ as input, 
\begin{equation}
 \phi_t=f_\theta(\operatorname{sg}(h_t)),\qquad
 r_t^{\mathrm{int}}=g_\omega(\phi_t,a_t,\phi_{t+1};\mathcal{M}_t),
 \label{eq:idm_interface}
\end{equation}
where $f_\theta$ is an optional intrinsic feature mapping, $\operatorname{sg}$ is a stopgrad operator, $r_t^{\mathrm{int}}$ is the intrinsic reward with parameters $\omega$, and $\mathcal{M}_t$ contains algorithm-specific state, such as episodic memory or a sampled skill. Predictors, density estimators, and discriminators that constitute a given intrinsic motivation all train on posterior trajectories (latent representations of real experience from replay) and their computed rewards train the exploration actor--critic on the imagined trajectories (from the prior). Hierarchical competency (i.e. skill-discovery) methods additionally condition the exploration policy on the skill sampled from a skill prior. We block IDM training gradients at the RSSM and stop gradients through intrinsic rewards during actor updates (Fig. \ref{fig:overview}). Each drive therefore shapes the world model through the experience it collects, without directly changing its representation loss.


\subsection{Overview and Taxonomy of Intrinsic Objectives}
Figure~\ref{fig:exploration_taxonomy} organizes intrinsic objectives by the computational quantity used to value experience: predictions for curiosity, the distribution of visited states for entropy, and the relationship between behavioral choices and their outcomes for competency. These families capture different premises for exploration, and some algorithms combine mechanisms from several families, so this taxonomy is meant as an organizational device. 

\paragraph{Curiosity} 
Curiosity rewards prediction error, uncertainty, or learning progress within an internal model of the world. In the simplest case, illustrated in Figure~\ref{fig:exploration_taxonomy}, an action-conditioned model predicts the next feature $\phi_{t+1}$, and the discrepancy between predicted and observed features rewards unexpected transitions. The prediction target need not be a future state, like in RND, where the targets are outputs of a fixed random network projection of the current observation \citep{burda2019rnd}. Other approaches instead use ensemble methods to estimate world-model information gain rather than prediction error \citep{pathak2019disagreement,shyam2019max}. 
\paragraph{Entropy}
Entropy objectives reward visits to sparsely sampled regions of the state or feature distribution estimated from past experience. Unlike curiosity, this encourages broad state coverage even when dynamics may already be predictable. APT, RE3, and Proto-RL estimate novelty through nearest-neighbor distances, where larger distances indicate lower local sample density \citep{liu2021apt,seo2021re3,yarats2021proto}. SMM instead uses a learned variational density proxy \citep{lee2019smm}.

\paragraph{Competency}
Competency objectives value the agent's ability to produce reliably distinguishable outcomes through its behavior. Their substrate is the statistical relationship between a controllable choice---an action or a skill---and the states, transitions, or trajectories that follow. Empowerment formalizes this relationship as the channel capacity from actions to future states, maximizing their mutual information over action distributions for a given initial state \citep{mohamed2015variational,zhao2021empowerment}. High empowerment therefore indicates that the agent has multiple distinguishable outcomes available through its actions. Skill-discovery methods apply a similar principle to temporally extended behavior by conditioning a policy on a random skill variable $z$ and then learning either a classifier to predict the skill from the policy's behavior or a regressor to predict the future dynamics under the skill~\citep{eysenbach2019diayn,gregor2017vic,achiam2018valor,sharma2020dads,laskin2022cic}. The discriminability illustrated in Figure~\ref{fig:exploration_taxonomy} therefore reflects behavioral diversity that the agent can identify via skills. 

\subsection{Environment Choices and Experiment Design}


\paragraph{Continuous control}
We evaluate all 25 tasks from the DeepMind Control (DMC) suite from proprioceptive observations using DreamerV3 \citep{tassa2018dmc}. Here, proprioception is used to emphasize that many of the tasks provide local joint states, not global variables. The vision modality uses a third-person camera, introducing the problem of inferring body position and movement from images. We therefore use proprioception for the full suite, then compare it with image-only observations on Walker Stand, Walk, and Run across all three backbones to measure the effects of modality on intrinsic motivation. DMC pretraining uses 1M environment steps. 

\paragraph{Open-ended exploration}
We use Craftax Classic, the JAX reimplementation of Crafter, as a complementary test of open-ended exploration from vision without strong dynamical structure to exploit (unlike DMC) \citep{matthews2024craftax,hafner2022crafter}. Its discrete actions express abstract operations such as gathering and crafting, while achievements (which constitute reward) vary from independent goals, like waking up, to those with compositional structure, like crafting a stone pickax. Progress therefore requires discovering resources and learning how to combine them over extended behavioral sequences. Craftax pretraining uses 5M environment steps.  

\begin{figure*}[t!]
  \centering
  \includegraphics[width=\textwidth,trim={0 8bp 0 8bp},clip]{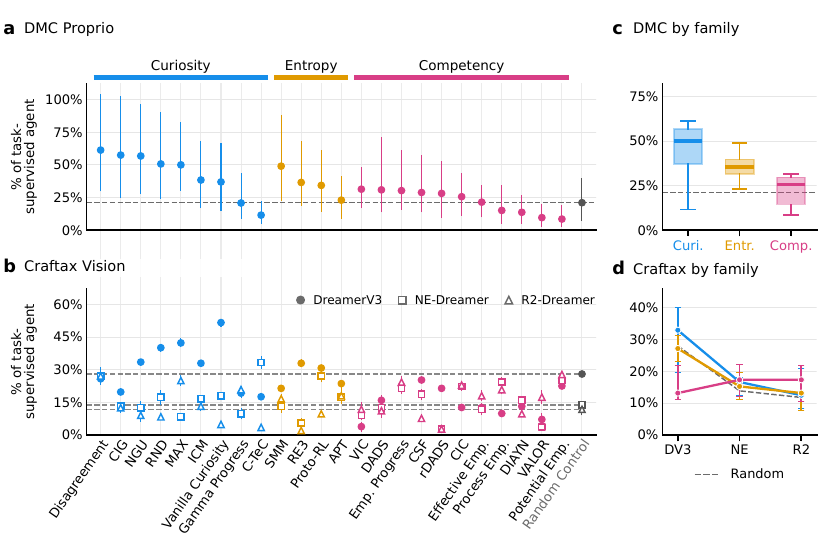}
  \caption{\textbf{Single-task zero-shot generalization.} Returns are normalized by the task-supervised reference. (a) DreamerV3 interquartile-mean (IQM) return across 25 proprioceptive DMC tasks after 1M exploration steps; error bars bootstrap tasks. (b) Craftax vision after 5M steps across three backbones. (c) DMC family distributions. (d) Craftax family medians and interquartile ranges across methods. Dashed lines denote Random Control.}
  \label{fig:zero_shot}
\end{figure*}

\paragraph{Multi-task transfer}
Walker-Stand, Walker-Walk, and Walker-Run share the same embodiment and environment dynamics with a natural compositional relationship, in that postural control supports walking and coordinated locomotion supports running. We use Stand as the source environment, then evaluate the shared model on all three rewards by training task-specific heads in imagination. For intrinsically motivated agents, ``Stand source'' identifies the environment in which data are collected, not the task driving exploration. The comparison with a Stand-supervised agent tests whether intrinsic exploration produces a more reusable model than experience selected for one task. For DreamerV3, we additionally test multitask transfer in RoboKitchen, a Franka Panda arm manipulation experiment in a fixed workspace with multiple distinct goals \citep{mendonca2021lexa}. Here, the source task is moving the kettle and the transfer tasks are interacting with a light-switch and sliding cabinet (Appendix Figure~\ref{fig:kitchen}).
\paragraph{Few-shot adaptation}
After intrinsic pretraining, we reinitialize the task actor, critic, and reward predictor, keep the world model and replay, and train the fresh heads for ~5k offline updates with the model fixed. We then unfreeze the model and collect task-directed experience---200K steps for Walker Stand-to-Run and 500K for Craftax---under the new task reward head, whose gradients now flow through the world model.  For a selected DreamerV3 subset, we compare 5K and 50K fresh-head updates on a frozen model with task heads retained from pretraining, which corresponds to 500K updates (Figure~\ref{fig:head_budget_average_vs_scratch}).

\begin{figure*}[t!]
  \centering
  \includegraphics[width=\textwidth,trim={0 2.5bp 0 8.5bp},clip]{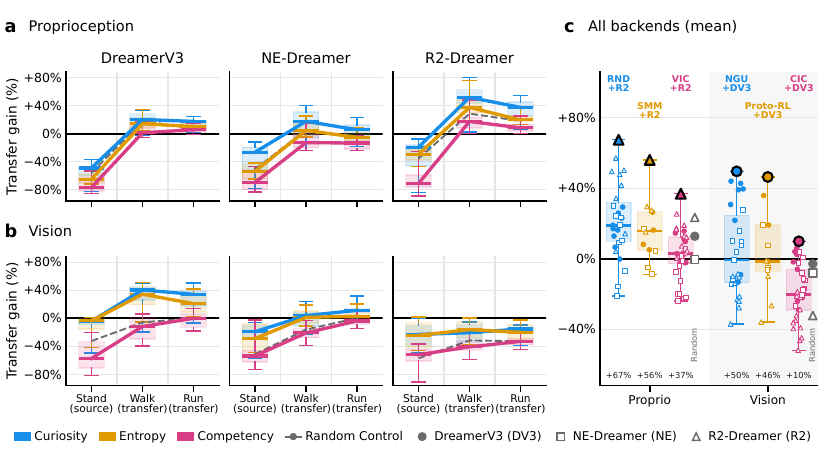}
  \caption{\textbf{Walker zero-shot transfer.} Performance on Stand, Walk, and Run after 1M pretraining steps in the Stand environment from (a) proprioception and (b) vision. Boxes span each family's interquartile range across methods, with medians and full-range whiskers. (c) Each method/backend pair's gain averaged over Walk and Run.}
  \label{fig:walker_family_gain_tasks}
\end{figure*}

\section{Results}
\label{sec:results}
\paragraph{Zero-shot Generalization}
Across both DMC and Craftax, we find that curiosity is the strongest intrinsic motivation for zero-shot control (Figure~\ref{fig:zero_shot}c). While several algorithms in the curiosity family achieve similar maximum performance (Disagreement, CIG, NGU) on DMC Proprio (Figure~\ref{fig:zero_shot}a), this ordering does not hold in Craftax. Interestingly, a simple prediction-error objective (Vanilla Curiosity) excels here, suggesting that a richer environment and feature space support a simpler self-supervised learning algorithm. On the other hand, many entropy and competency algorithms that exceeded random exploration in the DMC Proprio experiment now offer no pretraining benefit in Craftax Vision (Figure~\ref{fig:zero_shot}b). Although continuous and discrete control may contribute to these differences, we suspect competency algorithms especially struggle to scale to high-dimensional observations owing to their dependence on variational mutual information estimators. In DMC Proprio, many algorithms saturate task-supervised performance at 100\%,  whereas the best algorithm in Craftax Vision reaches 51\%. 

DreamerV3 generally outperforms the reconstruction-free world models in both DMC Proprio and Craftax Vision (Figure~\ref{fig:zero_shot}c-d). In Craftax, the most performant pretraining combination across both curiosity and entropy is with DreamerV3 (Figure~\ref{fig:zero_shot}b, d). Interestingly, competency methods pair better with reconstruction-free world models, with NE-Dreamer and R2-Dreamer both outperforming DreamerV3 competency pairings (Figure~\ref{fig:zero_shot}d). 

The walker transfer experiment tests the extent to which intrinsic motivation builds a model useful for zero-shot transfer to multiple distinct reward functions. We find that the strongest world-model backend depends on observation modality. R2-Dreamer can support strong transfer from proprioception, while DreamerV3 pairs particularly well with curiosity and entropy under vision (Figure~\ref{fig:walker_family_gain_tasks}a-c). 
Richer observations therefore do not uniformly improve intrinsic pretraining; instead, their value depends on how the world model represents them. We again observe that competency methods suffer a large performance deficit in vision-based transfer, whereas curiosity and entropy methods are quite robust. Despite these dependencies on observation modality, curiosity remains the single-most performant objective for pretraining from intrinsic motivation (Figure~\ref{fig:walker_family_gain_tasks}c). 

Thus, pretraining from intrinsic motivation can yield a zero-shot world model for proprioception-based control that is as, or nearly as, effective as a task-supervised agent (performance on individual environments is shown in Appendix Figure~\ref{fig:dmc_proprio_tasks}). These results hold for vision-based zero-shot transfer in Walker. This suggests that the performance difference in Craftax Vision arises from the nature of the environment itself rather than the observation modality alone. Unlike DMC, Craftax is not a physics-based control task where the world-model representation can exploit rich environment dynamics. Craftax requires more complex planning to achieve high reward due to the compositional achievement structure. This imposes strong constraints on the world model to maintain an episodic memory and represent difficult representation-reward contingencies. 

\paragraph{Few-shot Adaptation}
\begin{figure*}[t!]
  \centering
  \includegraphics[width=\textwidth,trim={0 2.5bp 0 6.5bp},clip]{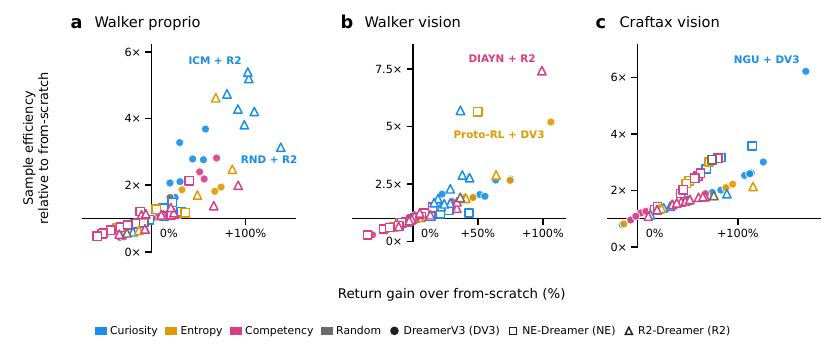}
  \caption{\textbf{Adaptation gain and sample efficiency.} All three backbones after 200K Stand-to-Run adaptation steps with (a) proprioception or (b) vision, and (c) 500K Craftax steps. Points to the right of zero achieve higher return than training from scratch at the same task-step budget. Points above $1\times$ require fewer task interactions to reach a matched return.
  The labeled algorithms identify the best combination on each axis.}
  \label{fig:gain_vs_multiplier}
\end{figure*}

In these experiments, we train all the task-relevant parts (actor-critic, reward head, and continuation head) from scratch, keeping only the pretrained world model and current replay buffer. This allows us to isolate the role of world-model adaptation from policy adaptation, unlike previous works that begin finetuning the world model from a pretrained policy \citep{sekar2020plan2explore}. This is an important design choice in order to understand what adaptation benefits we gain from the pretrained world model alone and its capacity to train a fresh task-specific policy on the fly. Figure~\ref{fig:gain_vs_multiplier} compares two different benefits of pretraining:  better performance after a fixed amount of task practice (x-axis), and fewer interactions needed to achieve a particular performance (y-axis). 

\begin{figure}[!t]
  \centering
  \begin{minipage}[t]{0.46\textwidth}
    \centering
    \vspace{0pt}
    \includegraphics[height=1.9in,width=\linewidth,keepaspectratio,trim={10.4bp 5bp 0 3bp},clip]{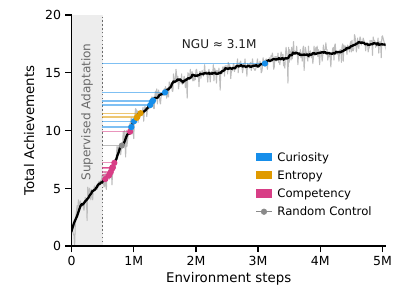}
    \caption[Finetuning sample complexity on Craftax.]{\textbf{Finetuning sample complexity on Craftax.} Horizontal lines locate DreamerV3 returns after 500K adaptation steps. Finetuning from NGU's world model is equivalent to training from scratch for 3.1M steps.}
    \label{fig:craftax_equivalent_steps}
  \end{minipage}\hfill
  \begin{minipage}[t]{0.46\textwidth}
    \centering
    \vspace{0pt}
    \includegraphics[height=1.9in,width=\linewidth,keepaspectratio,trim={0 5bp 0 2bp},clip]{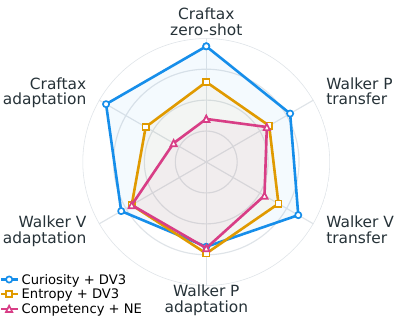}
    \caption{\textbf{Selected family profiles.} Top-four intrinsic motivations using each family's highest-mean backend across six settings, scaled (Appendix~\ref{sec:family_summaries}). P/V denote proprioception/vision.}
    \label{fig:performance_coverage_radar}
  \end{minipage}
\end{figure}

On Craftax, NGU with DreamerV3 attains the strongest result in both measures (Figure~\ref{fig:gain_vs_multiplier}c), reaching a final score of 15.8 versus 5.9 from scratch after 500K steps (a 167.8\% gain over training from scratch). Figure~\ref{fig:craftax_equivalent_steps} illustrates the task-interaction ratio explicitly, where horizontal lines locate adapted agents' returns on the learning curve of the supervised agent training from scratch. For NGU, this corresponds to approximately 3.11M steps, a $6.22\times$ improvement in sample complexity. 
Craftax vision is the only example in which a single pairing achieves the best pretraining performance in both measures. 
In Walker Proprio, ICM with R2-Dreamer yields the largest task-interaction ratio from proprioception, while  RND with R2-Dreamer achieves the best performance under a fixed budget (Figure~\ref{fig:gain_vs_multiplier}a). 
Thus, return gain and sample efficiency are positively associated but need not rank methods in absolute terms. Rather, one may be willing to sacrifice performance if environment interaction is expensive, preferring a method with greater sample efficiency.

The fact that NGU particularly excels in Craftax is consistent with our hypothesis that the nature of the compositional achievement structure benefits from a strong memory mechanism. NGU multiplicatively combines episodic novelty, measured by similarity to recent states in an inverse-dynamics embedding, with lifelong novelty, estimated by random network distillation (RND) prediction error \citep{badia2020ngu}. Thus, intrinsic motivation can implement memory that complements the world model’s recurrent state by tracking novelty at different timescales. Interestingly, an intrinsic motivation with two novely timescales was recently shown to reproduce exploration behavior and whole-brain activity in zebrafish, suggesting this mechanism plays an imporant role in general intelligence \citep{keller2026intrinsic}. 


\paragraph{When do bells and whistles help?}
Many algorithms within each family are variants of another, and it's not clear how these differences contribute to world-model pretraining. For example, ICM and C-TeC both introduce specific feature spaces for their intrinsic rewards, so we implement a Vanilla Curiosity that simply uses the world-model latent. Similarly, Disagreement, MAX, and CIG are all ensemble methods for estimating information gain, but each is a bit more complex than the last. Finally, NGU builds on RND by adding episodic novelty, and Proto-RL and our APT-style variant differ in both representation learning and feature sampling. In the same nine-setting comparison (Appendix Figure~\ref{fig:lineage}), Vanilla Curiosity (+17.7\%) exceeds ICM (+12.8\%) and C-TeC (+0.8\%), while Disagreement (+19.8\%) exceeds MAX (+18.0\%) and CIG (+10.7\%). However, episodic novelty in NGU improves RND's mean gain from +20.1\% to +23.8\%, and Proto-RL has a higher mean gain than our APT-style variant ($-1.7\%$ to +25.8\%). These results support retaining simple latent spaces where it matters, as added feature learning or more elaborate uncertainty estimates therefore do not guarantee better world models. For curiosity, the DreamerV3 world-model latent is already sufficient, whereas entropy estimation benefits from a replay representation. In the case of NGU, the additional mechanistic structure of a multiplicative interaction between short- and long-term memory shows significant potential for environments with complex reward structures (Figure~\ref{fig:gain_vs_multiplier}c). 

\Needspace{20\baselineskip}
\paragraph{More imagination or more experience?}
\begin{wrapfigure}[18]{r}{0.38\textwidth}
  \centering
  \vspace{\dimexpr-\intextsep-4pt\relax}
  \includegraphics[width=\linewidth,trim={0 2bp 0 9bp},clip]{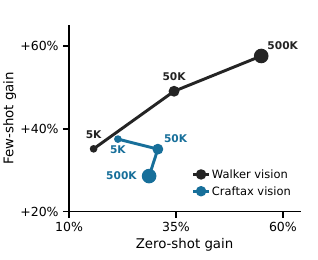}
  \caption[Head initialization and adaptation.]{\textbf{Imagination pretraining and adaptation.} Mean gains across top-seven intrinsic motivations. 5K/50K/500K denote offline updates of task heads in imagination.}
  \label{fig:head_budget_average_vs_scratch}
\end{wrapfigure}

Unlike previous work, we emphasize that the policy should be used deliberately as a tool to measure world-model adaptation, rather than just starting from a pretrained policy. We achieve this by independently varying the amount of offline training, where the policy learns entirely in a frozen world model, with online training, where the policy interacts with the real environment. 
We consider three offline training budgets of 5K, 50K, and 500K imagination steps and measure the agents' zero-shot and adaptation performance---the correlation between the two indicates the alignment between the world model and task. On Walker Vision, training in imagination improves both the zero-shot and adaptation performance monotonically (Figure~\ref{fig:head_budget_average_vs_scratch}). On the contrary, in Craftax Vision, training in imagination improves zero-shot performance up to 50K steps, but worsens adaptation performance. At 500K steps, it no longer improves zero-shot performance, but significantly worsens adaptation performance. These results expose a  weakness in DreamerV3's ability to model complex environments that contain long-term dependencies and compositional structure. The world model may accurately imagine familiar states while poorly representing transitions needed for further progress. Thus, more policy optimization can only exploit this local knowledge, and even become more committed to it to the detriment of adaptation. 

\section{Discussion}
\label{sec:discussion}

Animals are intrinsically motivated over their entire lifetime. What role does this play in intelligence? We hypothesize that intrinsic motivation serves as a developmental mechanism for interacting with the world in ways that are useful for building useful abstractions and internal models that enable quick adaptation to downstream goals. To understand the ingredients of intrinsic motivation and its relationship to developing world models, we propose a framework that starts from a general model-based reinforcement learning agent that supports three distinct world-model representation hypotheses. Then, we evaluate 24 different intrinsic motivation algorithms based on their capacity for zero-shot, few-shot, and multi-task generalization. 

We find that curiosity has the strongest overall family performance, but no single intrinsic drive is best across environments, modalities, and representations. This dependence favors designing intrinsic motivation and world-model representations jointly, especially considering a predictive representation as a common substrate of both modules. 
Further, simple intrinsic rewards, especially in the curiosity family, are often competitive with more sophisticated algorithms. Thus, intrinsic motivation can be expressed with simple mechanisms on top of a sufficiently strong world-model representation. 
We also show that offline planning does not always scale with task-directed finetuning. On the contrary, training more in imagination handicaps the policy's ability to adapt from environment experience even if it improves zero-shot control. If the role of a world model is to learn generalizable and robust abstractions, further pretraining in that world model should not leave the policy worse for wear. This scaling mismatch reflects a shortcoming in the ability to represent the dimensions of the environment relevant to the task, especially when the task imposes long timescale and compositional dependencies.

Taken together, our results show that curiosity-driven intrinsic motivation is a promising route to pretraining world models through autonomous interaction with the environment. However, we suspect further progress on challenging environments depends on at least two key ideas: \emph{persistent memory} that connects states separated in time, allowing earlier observations to be reinterpreted as an agent learns, and a \emph{representational bias toward affordances}, organizing knowledge around what an agent can achieve \citep{bastankhah2026learning}. 
Alternative world-model systems to Dreamer, such as Probabilistic Structure Integration (PSI) and Baby Zero-shot World Model (BabyZWM), provide useful steps in these directions that may support better intrinsic motivation \citep{kotar2025psi,aw2026zeroshot}.
PSI extracts structure from a probabilistic video model and incorporates it as additional conditioning inputs and prediction targets during further training, a potential mechanism by which an agent's experience could refine the model that guides subsequent exploration. 
BabyZWM learns a predictor that separates visual appearance from dynamics, then queries and combines its predictions to recover visual structure zero-shot. 
A world model whose learned physical knowledge can be accessed through counterfactual inference could enable intrinsic motivation towards experiences that reveal important physical relations between objects and their parts.
Together, these ideas may unlock the potential of building world models from intrinsic motivation that scales to the cognitive complexity and environmental richness of human ethology. 

\section{Limitations and Future Work}
Future work should extend this evaluation to more diverse environments and sensory modalities, especially in robotic manipulation environments like RoboCasa, which offers varied household scenes and compositional tasks \citep{nasiriany2024robocasa}. In addition to the task complexity, this also affords testing whether intrinsically motivated exploration in simulation builds world models that accelerate adaptation to real robots. 
Combining vision and proprioception with depth or tactile sensing would further reveal how complementary sensory information changes which intrinsic objectives support transferable representations~\citep{chung2026task,chung2026tactile}. 

Another opportunity is to investigate how intrinsically motivated agents can build on pretrained visual models, such as PSI \citep{kotar2025psi}. This would test whether autonomous interaction can adapt prior visual knowledge to unfamiliar dynamics and affordances in complex environments. A central question is whether these priors change which intrinsic objectives are most effective and reduce the experience needed for downstream control. 

Finally, we designed our experiments around a skill-agnostic world model in order to maintain a consistent architecture across algorithms and isolate the role of intrinsic motivation to collecting replay experience. Augmenting state-of-the-art world models to condition on skills would allow us to test the adaptation benefits of planning over the learned skills themselves, which may confer an advantage in long-horizon tasks. 

\subsection*{Acknowledgments}
We thank Andrea Bajcsy for helpful discussions and feedback at earlier phases of the project. RK and XP were supported in part by the Simons Collaboration in Ecological Neuroscience (SFI-AN-NC-SCN-00007276) and in part from funds provided by the National Science Foundation and by DoD OUSD (R \& E) under Cooperative Agreement PHY-2229929 (The NSF AI Institute for Artificial and Natural Intelligence, ARNI). D.K. was supported with funding and resources from Northrop Grumman, the United States Department of Defense High Performance Computing Modernization Program (HPCMP), and the State of Maryland.
A.N. was supported in part by the Burroughs Wellcome Fund (CASI award), Foresight Institute, and Protocol Labs.

\subsection*{AI use statement}
In this work, we used generative AI tools to assist with research ideation and execution, including implementation of aspects of the project using Codex. Additionally, we used generative AI tools to aid and polish writing, retrieve and discover related work, and create and format plots with ChatGPT and Claude. Codex-assisted implementation work was independently reviewed by the primary author. We take responsibility for the final content of this work, including text, claims or artifacts produced with the aid of generative AI.

\bibliography{iclr2027_conference}
\bibliographystyle{iclr2027_conference}
\clearpage
\appendix

\section{Related Work}
\label{sec:related}
\paragraph{Intrinsic motivation in cognitive and motor development}
The developmental role of intrinsic motivation has long been a focus of psychological theory. \citet{white1959competence} framed exploration and play as intrinsically motivated routes to developing competence. Infant motor exploration provides evidence for learning action--effect relations: four-month-olds show more limb-specific responses to an arm-controlled mobile than two-month-olds \citep{watanabe2006general}. Exploration is also structured by prior knowledge: infants investigate objects that violate physical expectations \citep{stahl2015unexpected}, while preschoolers preferentially play with causally ambiguous toys and generate disambiguating interventions \citep{schulz2007serious}. Motivated by these findings, our study tests which intrinsic objectives produce experience that builds predictive models useful for later control.

\paragraph{Intrinsic motivation as world-model learning}
The basic idea of computationalizing curiosity as the acquisition of knowledge about the environment has a long history in reinforcement learning, but methods operationalize that idea differently \citep{schmidhuber1991curious,oudeyer2007intrinsic,barto2013intrinsic}. Prediction error rewards surprising transitions, learning progress rewards improvements in prediction, and ensemble disagreement approximates uncertainty about learnable dynamics. Active world-model learning explicitly evaluates curiosity by the predictive model it builds \citep{kim2020progress}. Plan2Explore trains an exploration policy in imagination to maximize disagreement, then uses its world model for zero-shot and few-shot tasks \citep{sekar2020plan2explore}. These works establish the connection between intrinsic exploration and reusable models. Our contribution is a systematic comparison across a much larger collection of intrinsic objectives and multiple world-model representation strategies, rather than the first use of curiosity to learn a world model.

\paragraph{Unsupervised RL benchmarks and goal-conditioned control}
URLB unifies eight unsupervised algorithms across knowledge-based, data-based, and competence-based exploration, with reward-free pretraining followed by reward-directed fine-tuning on twelve DMC tasks \citep{laskin2021urlb}. It provides an important precedent for comparing families within one implementation. Our setting instead uses an explicit learned dynamics model for imagined control, evaluates task policies without task-directed collection, varies the world-model representation objective, and resets task heads to examine model-and-replay transfer. Related model-based precedents include the seven-objective DreamerV2 comparison of \citet{rajeswar2023mastering} and EUCLID, which jointly pretrains exploration and latent dynamics, using multiple heads to capture dynamics under different behaviors and selecting a head for downstream adaptation \citep{yuan2023euclid}. We extend these comparisons to 24 intrinsic objectives and three representation strategies. LEXA also learns a world model without task-directed exploration, coupling an explorer with a goal-conditioned achiever trained on imagined goals \citep{mendonca2021lexa}. We evaluate task reward heads rather than visual goal reaching, and cross the exploration objective with alternative world-model representations. 


\paragraph{Latent world models and joint-embedding prediction}
Joint-embedding predictive architectures (JEPA) learn by predicting in representation space. I-JEPA instantiates this idea by predicting representations of masked image regions from visible context \citep{assran2023ijepa}. Our reconstruction-free variants share this emphasis on latent representations, but couple it to action-conditioned recurrent dynamics for multi-step imagined control. TD-MPC2 provides a complementary control-oriented example: it learns decoder-free dynamics through joint-embedding, reward, and value prediction \citep{hansen2024tdmpc2}. In our setting, intrinsic exploration supplies the training experience and task-reward gradients are excluded from representation learning during pretraining, allowing us to evaluate how the resulting world model transfers to downstream control.

\section{Architecture and World-model Objectives}
\label{sec:architecture}
At time step $t$, let $x_t$ be the observation and $a_t$ the action. Dreamer's encoder and recurrent latent dynamics are
\begin{align}
 e_t &= E_\phi(x_t),
 &h_t &= F_\phi(h_{t-1},z_{t-1},a_{t-1}), \label{eq:rssm_rec}\\
 q_t &\doteq q_\phi(z_t\mid h_t,e_t),
 &p_t &\doteq p_\phi(z_t\mid h_t),\qquad s_t=(h_t,z_t). \label{eq:rssm_latent}
\end{align}
Here $E_\phi$ is the observation encoder, $e_t$ its embedding, and $F_\phi$ the recurrent transition. The parameter set $\phi$ collects world-model weights except for separately indexed modules; it is distinct from the intrinsic feature vector $\phi_t$ in Section~\ref{sec:methods}. The deterministic state $h_t$ carries recurrent memory; $q_t$ and $p_t$ are the posterior and prior distributions of the categorical state $z_t$. This state is sampled from the posterior during observation and the prior during imagination, and $s_t=(h_t,z_t)$ combines recurrent memory and stochastic state. The shared generative losses are
\begin{align}
 \mathcal{L}_{\mathrm{dyn},t}
 &=\max\!\left(1,\mathrm{KL}[\operatorname{sg}(q_t)\,\|\,p_t]\right), \nonumber\\
 \mathcal{L}_{\mathrm{rep},t}
 &=\max\!\left(1,\mathrm{KL}[q_t\,\|\,\operatorname{sg}(p_t)]\right), \label{eq:wm_kl}\\
 \mathcal{L}_{\mathrm{con},t}
 &=-\gamma c_t\log d_\phi(s_t)-(1-\gamma c_t)\log(1-d_\phi(s_t)), \nonumber\\
 \mathcal{L}_{\mathrm{base}}
 &=\mathbb{E}\!\left[-\log p_\omega(r_t\mid\bar{s}_t)+\mathcal{L}_{\mathrm{con},t}
 +\mathcal{L}_{\mathrm{dyn},t}+0.1\mathcal{L}_{\mathrm{rep},t}\right]. \label{eq:wm_base}
\end{align}
Here $r_t$ is the observed task reward, $c_t$ is the nontermination indicator, and $\gamma=1-1/333$. The reward distribution $p_\omega(r_t\mid\bar{s}_t)$ has separate parameters $\omega$. The continuation head $d_\phi(s_t)\in(0,1)$ is trained by binary cross-entropy against the soft target $\gamma c_t$; its prediction already includes discount. Time-limit truncation is distinguished from true termination. The operator $\operatorname{sg}$ is the identity in the forward pass and stops gradients, $\mathrm{KL}$ is Kullback--Leibler divergence, and $\log$ is the natural logarithm. The expectation averages over replay sequences and posterior samples. The KL terms train the prior and regularize the posterior, respectively; the one-nat floor applies after summing over categorical variables. During intrinsic pretraining, $\bar{s}_t=\operatorname{sg}(s_t)$: task rewards train a separate predictor but cannot shape the world model. Multi-task experiments optimize separate detached predictors for each evaluation reward. The replay-value representation loss is also disabled. Supervised training and online adaptation use $\bar{s}_t=s_t$ and additionally enable the replay-value objective with weight 0.3. Imagination actor--critic losses use detached model features in both modes; they do not directly update world-model parameters.

\paragraph{Representation losses}
DreamerV3 reconstructs observations using an image decoder or an MLP for symlog-transformed state observations. NE-Dreamer predicts the next encoder embedding from latent-action history using a two-layer, four-head causal transformer of width 256, with attention and prediction targets masked at episode boundaries. R2-Dreamer aligns the current latent state with the current encoder embedding through a bias-free linear projector. Let $U,V\in\mathbb{R}^{N\times d}$ contain paired predictor and target features, where $N$ is the number of valid pairs pooled over minibatch and time, and $d$ is their shared feature width. For R2-Dreamer, all same-step replay positions supply pairs. For NE-Dreamer, a pair is valid when $t$ is not an episode's last position and $t+1$ is not a reset; attention is causal and restricted to the same episode. Tildes denote independently centered columns divided by their population standard deviation plus $10^{-8}$ over the selected pairs. NE-Dreamer omits the alignment loss if fewer than two valid pairs remain. Define the cross-correlation matrix $C\in\mathbb{R}^{d\times d}$ and Barlow redundancy-reduction loss $\mathcal{B}$ by
\begin{equation}
 C=\widetilde U^\top\widetilde V/N,\qquad
 \mathcal{B}(U,V)=\sum_{i=1}^{d}(1-C_{ii})^2+5\!\times\!10^{-4}\!\sum_{\substack{1\leq i,j\leq d\\i\ne j}}C_{ij}^2. \label{eq:barlow}
\end{equation}
The superscript $\top$ denotes matrix transpose, and $C_{ij}$ is the correlation between feature $i$ of $U$ and feature $j$ of $V$. The diagonal term aligns corresponding dimensions, while the off-diagonal term discourages redundancy. The three world-model objectives during intrinsic pretraining are
\begin{align}
 \mathcal{L}^{(\mathrm{DV3})}_{\mathrm{WM}}
 &=\mathcal{L}_{\mathrm{base}}+\mathbb{E}[-\log p_\phi(x_t\mid s_t)], \label{eq:dv3_loss}\\
 \mathcal{L}^{(\mathrm{NE})}_{\mathrm{WM}}
 &=\mathcal{L}_{\mathrm{base}}+\mathcal{B}\!\left(T_\xi(s_{\leq t},a_{\leq t}),\operatorname{sg}(e_{t+1})\right), \label{eq:ne_loss}\\
 \mathcal{L}^{(\mathrm{R2})}_{\mathrm{WM}}
 &=\mathcal{L}_{\mathrm{base}}+0.05\,\mathcal{B}\!\left(P_\phi(s_t),\operatorname{sg}(e_t)\right). \label{eq:r2_loss}
\end{align}
The arguments of $\mathcal{B}$ denote matrices formed by stacking the valid batch--time pairs. Here $p_\phi(x_t\mid s_t)$ is the observation decoder's predictive distribution, $T_\xi$ is the causal transformer predictor with parameters $\xi$, and $P_\phi$ is the linear latent-to-embedding projector. The histories $s_{\leq t}$ and $a_{\leq t}$ include states and actions through step $t$ within the episode. The superscripts DV3, NE, and R2 identify DreamerV3, NE-Dreamer, and R2-Dreamer; WM denotes world model. NE-Dreamer uses a one-step prediction horizon and no transformer dropout. Both targets come from the current encoder with stopped gradients, rather than an exponential-moving-average target encoder. The encoder still learns through the posterior state used by the predictor and the other active model losses. Because these objectives retain different auxiliary architectures and scales, the comparison tests complete representation strategies within the shared Dreamer control framework.

\paragraph{Policy learning}
From replay posterior states, an actor $\pi_\theta(a_t\mid s_t)$ generates imagined trajectories through the RSSM prior. The current critic's value estimate $v_\psi$ supplies the bootstrap values in the $\lambda$-return:
\begin{equation}
 R_t^\lambda=r_{t+1}+d_\phi(s_{t+1})\left[(1-\lambda)v_\psi(s_{t+1})+\lambda R_{t+1}^\lambda\right]. \label{eq:lambda_return}
\end{equation}
Here $\theta$ and $\psi$ are actor and critic parameters, and $\lambda=0.95$ mixes one-step and longer-horizon targets. The reward $r_{t+1}$ is the intrinsic or predicted task reward for the transition from $s_t$ to $s_{t+1}$. At the final imagined state, $R_H^\lambda=v_\psi(s_H)$, where $H=15$. Because $d_\phi$ predicts discounted continuation, no additional $\gamma$ factor appears. The actor uses a score-function objective with stopped advantages, normalized by a moving estimate of the 95th--5th percentile return range with a lower bound of one, and entropy regularization. Actor and critic losses are weighted by cumulative predicted continuation. The critic fits stopped return targets and includes a unit-weight regularizer toward a stopped prediction from a slow-moving value network; that network supplies the regularization target, not the bootstrap values. Exploration and task heads use intrinsic and predicted task reward, respectively. Their imagination losses leave the shared model parameters unchanged.

\begin{table}[t!]
\centering
\caption{\textbf{Shared host configurations.} Model names identify the nominal host presets; adding an IDM or changing the representation predictor changes total parameter count. DMC/Walker and RoboKitchen use the small preset; Craftax uses the large preset.}
\label{tab:architecture}
\begin{tabular}{lrr}
\toprule
Setting & Small (12M preset) & Large (200M preset)\\
\midrule
Deterministic RSSM state & 2,048 & 8,192\\
RSSM hidden width & 256 & 1,024\\
Stochastic variables & 32 & 32\\
Categories per variable & 16 & 64\\
MLP width & 256 & 1,024\\
CNN base depth & 16 & 64\\
Imagination horizon & 15 & 15\\
Batch size $\times$ sequence length & $16\times64$ & $16\times64$\\
Replay capacity & 500,000 & 500,000\\
\bottomrule
\end{tabular}
\end{table}

The shared optimizer applies adaptive gradient clipping, RMS normalization, and then momentum (LaProp-style). Defaults are learning rate $4\times10^{-5}$, momentum coefficient $\beta_1=0.9$, squared-gradient decay $\beta_2=0.999$, numerical offset $10^{-20}$, clipping threshold 0.3, and 1,000 warmup updates. The discount horizon is 333, $\lambda=0.95$, actor entropy weight $3\times10^{-4}$, and slow-value update fraction 0.02. Walker uses 16 training environments, action repeat 1, and a replay training ratio of 512; Craftax uses a ratio of 32. Images are $64\times64$. These host settings are shared within comparisons; intrinsic modules retain their own source-informed settings, as described below.

\section{Intrinsic Drive Implementations}
\label{sec:idm}

\paragraph{Shared interface}
Each IDM evaluates the deterministic RSSM state $h_t$, either directly or through its own feature mapping. The exploration policy still receives the full state $s_t=(h_t,z_t)$. Auxiliary predictors, discriminators, and feature learners are trained on posterior trajectories inferred from real replay, while intrinsic rewards are computed on prior trajectories generated in imagination. Auxiliary losses stop at the RSSM, and intrinsic rewards are treated as fixed targets during behavior learning. Thus, each drive changes the data collected by the agent without adding a direct representation loss to the world model.

\paragraph{Policy learning and reward scaling}
We replace each source algorithm's policy learner with Dreamer's imagined actor--critic. The exploration policy optimizes intrinsic reward only, while separate task heads learn downstream control. Actor--critic settings, including Dreamer's return normalization, are shared across IDMs (Appendix~\ref{sec:architecture}). Intrinsic reward transformations, scaling factors, and normalization remain specific to each objective.

\paragraph{Skills and temporal context}
Skill-discovery methods concatenate a sampled skill to the exploration actor's input. The world model continues to predict from latent state and action, keeping its dynamics architecture the same across drives. During collection, skills persist for an episode or for the duration specified by the skill-horizon in the tables. During imagination, a fresh skill is sampled and held fixed for the full 15-step rollout. VIC and VALOR use 15-step options to align their terminal skill rewards with this horizon. Task policies do not receive exploration skills.

\paragraph{Novelty memory}
Memory-based novelty also requires an adaptation to imagination. For example, APT and RE3 query persistent memories of replay features, while Proto-RL queries a candidate queue updated from replay. Imagined states are evaluated against these memories without being inserted as new experience. NGU instead compares each state with earlier states in the supplied trajectory, so its episodic novelty during imagination is limited to the rollout context. Tables~\ref{tab:curiosity}--\ref{tab:competency} summarize the implemented objectives, their auxiliary architectures and settings, and substantive departures from the source formulations.


\begin{table}[p]
\centering
\caption{\textbf{Curiosity objectives (9).} All operate through the common Dreamer interface.}
\label{tab:curiosity}
\begin{tabular}{>{\raggedright\arraybackslash}p{0.24\textwidth}>{\raggedright\arraybackslash}p{0.69\textwidth}}
\toprule Method & Objective and defining implementation choices\\\midrule
Vanilla Curiosity \citep{pathak2017curiosity} & Local prediction-error baseline using the Euclidean error of an action-conditioned next-RSSM-feature predictor. Unlike ICM, it has no learned intrinsic feature projector; its auxiliary inverse head does not train the shared world-model representation.\\[4pt]
ICM \citep{pathak2017curiosity} & Forward prediction error in features learned jointly with inverse dynamics. A latent MLP replaces the original visual encoder and operates above stopped RSSM features; forward-loss weight 0.2 and reward coefficient 0.01.\\[4pt]
Disagreement \citep{pathak2019disagreement,sekar2020plan2explore} & Mean ensemble variance across five bootstrapped action-conditioned next-feature predictors. We retain the Disagreement/Plan2Explore novelty principle in RSSM space; ReLU predictors concatenate actions at input, with reward scale 10,000 and no log or internal reward normalization.\\[4pt]
CIG \citep{joseph2026cig} & Rollout-prefix-conditioned information-gain surrogate: per-step rewards are log-determinant increments of a regularized ensemble-disagreement Gram matrix. We use RSSM targets and five independently initialized predictors trained on the same replay transitions without bootstrapping; four 400-unit ELU layers, ridge multiplier 1, and 0.99 aleatoric/reward moving averages.\\[4pt]
MAX \citep{shyam2019max} & Jensen--R\'enyi divergence among probabilistic next-feature predictions, using MAX's ensemble novelty principle within Dreamer imagination. The plotted fast preset uses five predictors with two 256-unit ReLU layers and log-variance bounds $[-10,2]$, replacing the larger source ensemble.\\[4pt]
RND \citep{burda2019rnd} & Prediction error against a fixed random 512-dimensional target mapping. MLP target/predictor networks operate on RSSM features, and rewards are divided by a running mean; the source visual encoder and PPO normalization pipeline are replaced.\\[4pt]
NGU \citep{badia2020ngu} & Inverse-dynamics neighbor novelty multiplied by clipped lifelong RND novelty. The 32-dimensional feature memory is restricted to the supplied rollout/replay sequence, and one intrinsic Dreamer policy replaces the source policy family. Settings: 10 nearest neighbors, overall scale 0.3, kernel $\epsilon=10^{-4}$, cluster distance 0.008, pseudocount 0.001, similarity cutoff 8, and lifelong clip 5.\\[4pt]
$\gamma$-Progress \citep{kim2020progress} & Signed reduction in next-feature prediction error relative to a slow parameter copy (decay 0.9995). We retain the progress-curiosity signal while predicting stopped RSSM features and training the exploration policy in Dreamer imagination.\\[4pt]
C-TeC \citep{mohamed2026ctec} & L1 distance between a state--action embedding and a sampled future-state embedding, with 64-dimensional normalized features trained by InfoNCE and learned temperature initialized at 1. Our RSSM-based adaptation truncates discount-0.99 geometric future sampling to the available replay or imagined trajectory.\\
\bottomrule
\end{tabular}
\end{table}

\begin{table}[t!]
\centering
\caption{\textbf{Entropy objectives (4).} Feature geometry and density estimation differ across methods.}
\label{tab:entropy}
\begin{tabular}{>{\raggedright\arraybackslash}p{0.24\textwidth}>{\raggedright\arraybackslash}p{0.69\textwidth}}
\toprule Method & Objective and defining implementation choices\\\midrule
APT \citep{liu2021apt} & APT-style particle-entropy reward from 16 nearest neighbors in a normalized 256-dimensional projection, with clipping 0.0005. This local variant trains its projector to increase neighbor novelty instead of using the source contrastive loss; persistent feature memory and running-mean normalization are additional adaptations.\\[4pt]
RE3 \citep{seo2021re3} & Novelty from the distance to the 53rd-nearest neighbor in a fixed random embedding. Our RSSM-based recipe uses a 50-dimensional MLP, coefficient 0.1, and RMS-normalized unlogged Euclidean distances against persistent replay-feature memory.\\[4pt]
Proto-RL \citep{yarats2021proto} & Prototype-assignment prediction with novelty measured by distance to the third-nearest candidate in a feature queue. MLPs above RSSM states replace the source visual representation learner; 128-dimensional features, 512 prototypes, four candidates per prototype, temperature 0.1, three Sinkhorn iterations, and target momentum 0.95.\\[4pt]
SMM \citep{lee2019smm} & Single-policy SMM-inspired novelty: negative reconstruction log likelihood plus 0.5 times the posterior KL, with the decoder evaluated at the posterior mean. The VAE uses latent width 128 and two 150-unit ReLU layers. This is a novelty proxy, not an exact negative log-density or the source mixture-policy SM4 agent.\\
\bottomrule
\end{tabular}
\end{table}

\begin{table}[p]
\centering
\caption{\textbf{Competency objectives (11).} Task actors do not inherit the exploration skill input.}
\label{tab:competency}
\begin{tabular}{>{\raggedright\arraybackslash}p{0.24\textwidth}>{\raggedright\arraybackslash}p{0.69\textwidth}}
\toprule Method & Objective and defining implementation choices\\\midrule
DIAYN \citep{eysenbach2019diayn} & State-based skill discrimination, rewarding the sampled skill's discriminator log likelihood minus its log prior probability. We use 50 uniform categorical skills held for an episode, two 300-unit discriminator layers, and batch size 128. Dreamer's imagined actor--critic and entropy regularizer replace the source SAC learner.\\[4pt]
VIC \citep{gregor2017vic} & Initial/final-state option discrimination, with terminal reward equal to the selected option's decoder log likelihood minus its log prior probability. We instantiate VIC's option-information objective with 12 uniform categorical options and fixed 15-step duration matched to imagination.\\[4pt]
VALOR \citep{achiam2018valor} & Trajectory-based option decoding from 10 ordered differences between 11 equally spaced observations. We retain the trajectory-information objective but use an MLP decoder and 12 categorical 15-step options, replacing the source bidirectional recurrent decoder and growing option curriculum.\\[4pt]
DADS \citep{sharma2020dads} & Skill-conditioned next-feature log likelihood minus its log-mean likelihood under 100 prior-sampled alternatives. Skills are uniform on $[-1,1]^2$ with collection horizon 200 and two 512-unit predictor layers; a diagonal Gaussian over RSSM features replaces the source mixture-density model.\\[4pt]
rDADS \citep{sharma2020dads} & Local reverse surrogate: skill log likelihood conditioned on the transition minus that conditioned on the current state, using two learned Gaussian skill densities. It shares DADS's two-dimensional box prior and collection horizon 200, but changes the prediction direction and density estimators.\\[4pt]
CIC \citep{laskin2022cic} & Contrastive skill--transition representation learning with particle-neighbor novelty. The local module uses RSSM-transition MLPs and persistent feature memory; 64-dimensional skills on $[0,1]^{64}$, horizon 50, 64-dimensional features, two 1,024-unit layers, temperature 0.5, 16 nearest neighbors, and batch size 1,024.\\[4pt]
CSF \citep{zheng2025csf} & CSF's contrastive representation objective and feature-displacement/skill inner-product reward. Dreamer replaces the source successor-feature critic; four-dimensional unit-sphere skills held for an episode, two 1,024-unit projection layers, negative-term weight 5, and 256 prior-sampled negatives.\\[4pt]
Effective Empowerment \citep{mohamed2015variational} & Local one-step action-information surrogate: next-feature log likelihood given the current feature and action minus that given the current feature alone, using separate Gaussian predictors. This forward-density ratio differs from the cited inverse-action variational estimator and has no inner channel-capacity maximization or persistent skill input.\\[4pt]
Empowerment Potential \citep{zhao2021empowerment} & One-step capacity of a learned latent Gaussian channel; latent width 64, power 1, and 32 water-filling iterations. Our adaptation uses raw action vectors and a state-dependent drift in place of the source action-sequence encoder.\\[4pt]
Empowerment Process \citep{zhao2021empowerment} & Local Gaussian information surrogate under the exploration policy's action covariance, estimated from 16 action samples. It does not maximize capacity over action distributions; accumulated one-step rewards do not establish full feedback capacity.\\[4pt]
Empowerment Progress \citep{zhao2021empowerment,kim2020progress} & Local combination of Gaussian-channel empowerment and learning progress: current estimated capacity minus that of a slow copy (decay 0.9995). This signed score measures changes in estimated capacity, not a guaranteed increase in true controllability.\\
\bottomrule
\end{tabular}
\end{table}

\paragraph{Random Control}
For continuous actions, at each step we sample each component independently from a standard Gaussian, clip it to $[-1,1]$, add independent standard Gaussian noise, and clip again before rescaling to the action bounds. For discrete actions, we sample uniformly from the available actions at each step.

\clearpage
\section{Evaluation Details}
\label{sec:evaluation}
\subsection{Return normalization}

Let $J$ denote mean episodic evaluation return, with subscripts specifying the method and setting. Walker zero-shot transfer gain, $G^{\mathrm{transfer}}$, is the return difference expressed as a percentage of the task-supervised reference:
\begin{equation}
 G^{\mathrm{transfer}}_{m,b,o,k}
 =100\,\frac{J_{m,b,o,k}-J_{\mathrm{Stand},b,o,k}}
 {J_{\mathrm{task},\mathrm{DV3},\mathrm{proprio},k}}.
\end{equation}
Here $m$ is the intrinsic method, $b$ the backend, $o$ the observation modality, and $k$ the evaluation task. Thus $J_{m,b,o,k}$ is the method's evaluation return, $J_{\mathrm{Stand},b,o,k}$ the return supported by a Stand-supervised collector, and $J_{\mathrm{task},\mathrm{DV3},\mathrm{proprio},k}$ the task-supervised DreamerV3 proprioceptive return on task $k$. The Stand reference trains downstream task heads; it is not a task-supervised oracle on Walk or Run. Denominators are the task-supervised DreamerV3 proprioceptive scores (approximately 956, 944, and 553 for Stand, Walk, and Run). Positive gain therefore means improved use of experience relative to a Stand-specialized collector, not necessarily performance exceeding task-specific supervised training.

\begin{figure}[ht!]
  \centering
  \includegraphics[width=0.65\textwidth]{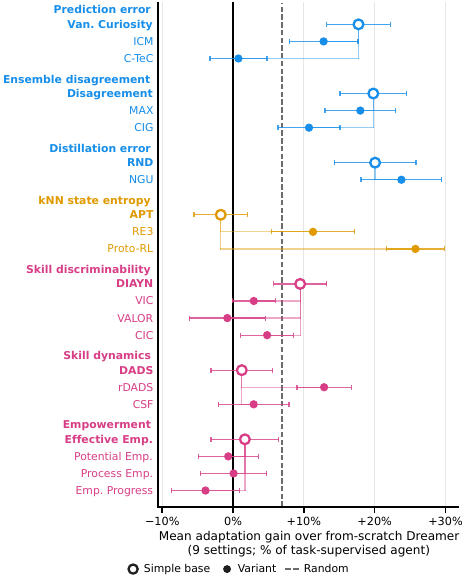}
  \caption{\textbf{Simple intrinsic rewards and related extensions.} Each group compares a simple objective (open circle) with methods adding feature learning, memory, density estimation, or a different mutual-information estimator (filled circles). Values average normalized adaptation gain across nine settings: Walker proprioception, Walker vision, and Craftax, each with three backends. Error bars are SEM across settings; the dashed line is Random Control. Groupings indicate conceptual relationships; variation in features, estimators, and hyperparameters prevents attributing differences to complexity alone.}
  \label{fig:lineage}
\end{figure}

For adaptation, let $B$ be the number of task-directed environment steps. The adaptation gain $G^{\mathrm{adapt}}$ used in Figures~\ref{fig:head_budget_average_vs_scratch} and \ref{fig:lineage} is normalized by task-supervised return:
\begin{equation}
 G^{\mathrm{adapt}}_{m,b,o}(B)
 =100\,\frac{J_{m,b,o}(B)-J_{\mathrm{scratch},b,o}(B)}{J_{\mathrm{task}}}.
\end{equation}
Here $J_{m,b,o}(B)$ and $J_{\mathrm{scratch},b,o}(B)$ are the adapted and from-scratch mean returns after $B$ task steps, and $J_{\mathrm{task}}$ is the task-supervised normalization score for the setting; the fixed evaluation task is suppressed in the subscripts. Walker uses a common Run normalizer of approximately 553; Craftax uses backend-specific supervised normalizers as above. Figure~\ref{fig:gain_vs_multiplier} reports $100[J_{m,b,o}(B)-J_{\mathrm{scratch},b,o}(B)]/J_{\mathrm{scratch},b,o}(B)$, the percentage gain over the matched from-scratch return. Thus NGU's Craftax gain is $100(15.8-5.9)/5.9\approx167.8\%$. DreamerV3 Craftax endpoint gains use matched greedy evaluations; the alternative-backend references use the saved supervised learning curves. The head-initialization figure subtracts the scratch score at step zero on its horizontal axis and at budget $B$ on its vertical axis. Its diagonal denotes equal normalized advantage at those two times.

\subsection{Task-interaction efficiency}
Let $f$ be a running-best mean-evaluation learning curve indexed by task-directed environment steps, $y$ a target score, and $\tau_f(y)$ the first step count at which $f$ reaches $y$, linearly interpolated between evaluations. The subscripts $\mathrm{adapt}$ and $\mathrm{scratch}$ identify the adapted and from-scratch curves; $J_{\mathrm{adapt}}(B)$ and $J_{\mathrm{scratch}}(B)$ are their returns at budget $B$, with method and setting indices suppressed. For Walker, with $B=200$K, the dimensionless task-interaction multiplier $M$ is
\begin{equation}
 M=\begin{cases}
 B/\tau_{\mathrm{adapt}}(J_{\mathrm{scratch}}(B)),
 &\text{if adaptation reaches the scratch endpoint},\\
 \tau_{\mathrm{scratch}}(J_{\mathrm{adapt}}(B))/B,
 &\text{otherwise}.
 \end{cases}
\end{equation}
A crossing is unavailable if the target is not reached within the recorded curve; a target already attained at the first evaluation is assigned that evaluation's step count. For Craftax, adaptation is evaluated more coarsely, so all multipliers use $\tau_{\mathrm{scratch}}(J_{\mathrm{adapt}}(500\mathrm{K}))/500\mathrm{K}$. The DreamerV3 reference is smoothed over 150K steps before taking its running maximum. These are related measures of task-interaction savings, but Walker and Craftax do not use identical matching rules. Neither includes pretraining interactions, offline head updates, or wall-clock computation.

Two reference differences further limit exact efficiency comparisons. DreamerV3 Walker proprioception uses an MJX from-scratch reference at 200,064 steps, while adaptation uses native \texttt{dm\_control} at 200,000; other Walker references use native \texttt{dm\_control}. DreamerV3 Craftax endpoint gain uses greedy evaluation, while its efficiency estimate reads the saved sampled-policy supervised learning curve. The reported multipliers should therefore be interpreted as descriptive estimates under these reference protocols. 

\subsection{Family summaries}
\label{sec:family_summaries}
The radar uses six settings shared by all backends: Craftax zero-shot control, Walker proprioceptive and visual transfer (each averaged over Walk and Run), Walker proprioceptive and visual adaptation, and Craftax adaptation. The 25-task DMC suite is excluded because it is available only for DreamerV3. Within each backend/setting, we rank all 24 intrinsic methods, assign average ranks to ties, and map rank $r$ to $100(r-1)/23$. Here $r$ runs from 1 (worst) to 24 (best), so the normalized rank spans 0--100; it is distinct from the time-indexed reward $r_t$. We select the four highest mean-ranked members of each family and backend using those same six settings. We then select one backend per family by the highest six-setting mean of that top-four profile and plot its fixed set of four methods on every spoke. This selects DreamerV3 for curiosity (83.15) and entropy (65.13), and NE-Dreamer for competency (52.54). These are the strongest relative family profiles under within-backend ranks, not necessarily the backends with the highest absolute returns. Exact ties are broken by algorithm name when selecting methods and by backend name when selecting backends. Curiosity and competency have nine and eleven eligible methods; entropy uses all four. Both selection stages use the displayed data and are descriptive rather than held-out validation. The equal setting weights give Walker four axes and Craftax two. Radar area depends on axis order and is not an additional performance metric.

\clearpage
\section{Supplementary Control Results}
\label{sec:per_method_tables}
\begin{figure*}[ht!]
  \centering
  \includegraphics[width=\textwidth]{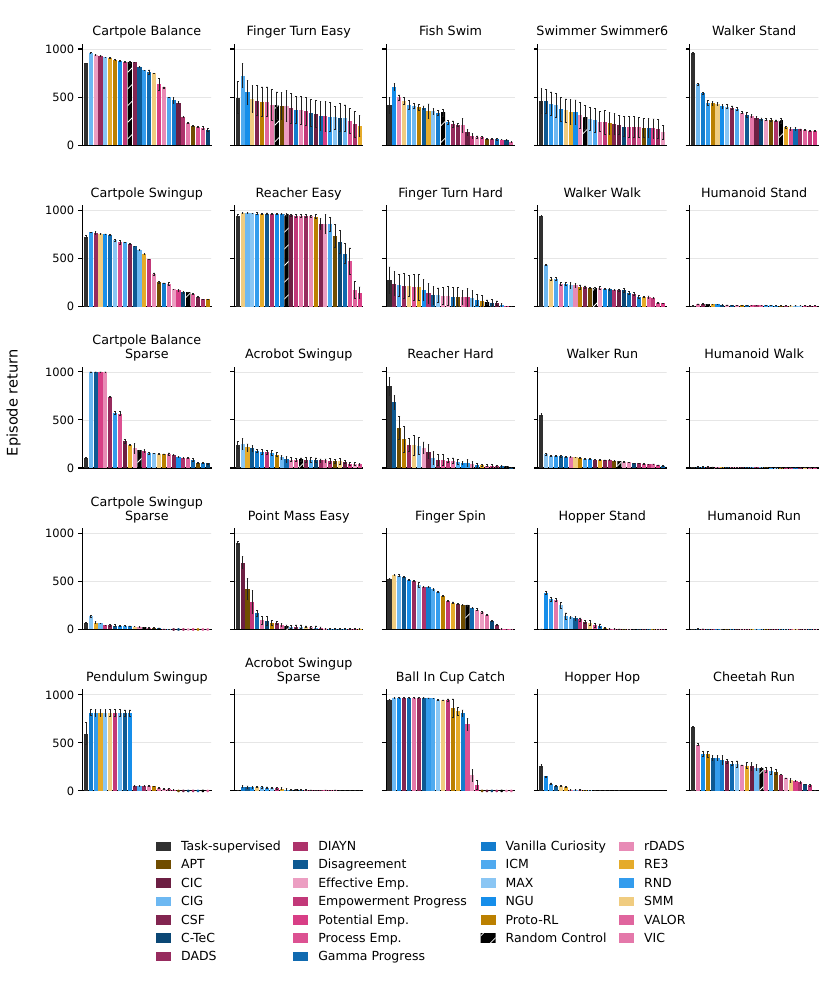}
  \caption{\textbf{Per-task zero-shot control across the 25 proprioceptive DMC tasks.} Bars show raw mean episodic return after 1M environment steps with DreamerV3; error bars denote evaluation-episode SEM across seeds. The task-supervised reference is first in each panel; the remaining 24 intrinsic methods and Random Control are sorted by return. Hatched black bars denote Random Control, and baseline dots mark zero returns. Colors identify intrinsic families and shades distinguish methods.}
  \label{fig:dmc_proprio_tasks}
\end{figure*}

\clearpage
\clearpage
\section{Manipulation Transfer in RoboKitchen}
\label{sec:kitchen}
\begin{figure*}[ht!]
  \centering
  \includegraphics[width=\textwidth]{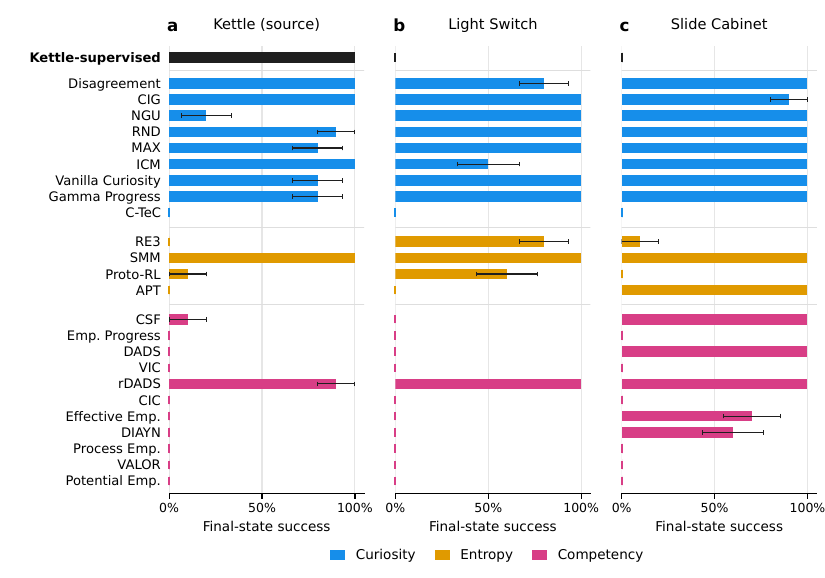}
  \caption{\textbf{Zero-shot manipulation transfer.} Strict final-state success after 1M environment steps with DreamerV3 (12M preset) on three of six RoboKitchen tasks: (a) Kettle (source), (b) Light Switch, and (c) Slide Cabinet. All 24 intrinsic methods and the Kettle-supervised reference are shown. The supervised reference (black) uses its source policy in (a) and detached task heads in (b,c); intrinsic methods use detached heads throughout. Colors denote intrinsic families. Bars show means and whiskers SEM across seeds; zero markers denote measured failures. Among the three omitted tasks, RND reaches 100\% Microwave success, no intrinsic method solves Hinge Cabinet, and only Proto-RL has nonzero Bottom Burner success (20\%). The displayed transfer therefore does not establish full task coverage. Historical and expanded method cohorts use different frozen source versions, so this comparison provides descriptive evidence rather than a precise algorithm ranking.}
  \label{fig:kitchen}
\end{figure*}

\end{document}